\documentclass[conference]{IEEEtran}
\IEEEoverridecommandlockouts
\usepackage{cite}
\usepackage{amsmath,amssymb,amsfonts}
\usepackage{algorithmic}
\usepackage{graphicx}
\usepackage{textcomp}
\usepackage{xcolor}
\def\BibTeX{{\rm B\kern-.05em{\sc i\kern-.025em b}\kern-.08em
    T\kern-.1667em\lower.7ex\hbox{E}\kern-.125emX}}

\usepackage{graphicx}
\usepackage{svg}
\usepackage{caption}
\usepackage{subfig}

\usepackage{amssymb}
\usepackage{amsmath}
\usepackage{mathtools}
\usepackage{dsfont}
\usepackage{cite}
\usepackage{stfloats}
\usepackage{psfrag}
\usepackage[mathscr]{euscript}
\usepackage[ruled, linesnumbered]{algorithm2e}
\usepackage{acronym}  
\usepackage{booktabs}
\usepackage{bbm}

\usepackage{multirow}

\usepackage{multicol}

\usepackage{xcolor}
\newcommand{\algrule}[1][.2pt]{\par\vskip.5\baselineskip\hrule height #1\par\vskip.5\baselineskip}

\usepackage{float}
\usepackage{balance}

\acrodef{CCDF}{complementary cumulative distribution function}
\acrodef{CF}{characteristic function}
\acrodef{PPP}{Poisson point processe}
\acrodef{RV}{random variable}
\acrodef{i.i.d.}{independent and identically distributed}
\acrodef{PDF}{probability distribution function}
\acrodef{CDF}{cumulative distribution function}
\acrodef{ch.f.}{characteristic function}
\acrodef{AWGN}{additive white Gaussian noise}
\acrodef{SNR}{signal-to-noise ratio}
\acrodef{LRT}{likelihood ratio test}
\acrodef{DRT}{distance ratio test}
\acrodef{GLRT}{generalized likelihood ratio test}
\acrodef{CRLB}{Cram\'{e}r-Rao lower bound}
\acrodef{CRB}{Cram\'{e}r-Rao bound}
\acrodef{ZZLB}{Ziv-Zakai lower bound}
\acrodef{ZZB}{Ziv-Zakai bound}
\acrodef{LOS}{line-of-sight}
\acrodef{ToF}{time-of-flight}
\acrodef{NLOS}{non-line-of-sight}
\acrodef{GDOP}{geometric dilution of precision}
\acrodef{GPS}{Global Positioning System}
\acrodef{FIM}{Fisher information matrix}
\acrodef{PEB}{position error bound}
\acrodef{SPEB}{squared position error bound}
\acrodef{TOA}{time-of-arrival}
\acrodef{TOF}{time-of-flight}
\acrodef{WSN}{wireless sensor network}
\acrodef{MAC}{medium access control}
\acrodef{RSS}{received signal strength}
\acrodef{WAF}{wall attenuation factor}
\acrodef{TDOA}{time difference-of-arrival}
\acrodef{RF}{radiofrequency}
\acrodef{RTT}{round-trip time}
\acrodef{AOA}{angle-of-arrival}
\acrodef{MF}{matched filter}
\acrodef{ED}{energy detector}
\acrodef{ML}{maximum likelihood}
\acrodef{MSE}{mean-square error}
\acrodef{RMSE}{root-mean-square error}
\acrodef{LEO}{localization error outage}
\acrodef{ppm}{part-per-million}
\acrodef{ACK}{acknowledge}
\acrodef{UWB}{Ultrawide bandwidth}
\acrodef{TNR}{threshold-to-noise ratio}
\acrodef{LS}{least squares}
\acrodef{IR-UWB}{impulse radio UWB}
\acrodef{FCC}{Federal Communications Commission}
\acrodef{TH}{time-hopping}
\acrodef{PPM}{pulse position modulation}
\acrodef{MUI}{multi-user interference}
\acrodef{PDP}{power delay profile}
\acrodef{BPZF}{band-pass zonal filter}
\acrodef{SIR}{signal-to-interference ratio}
\acrodef{SINR}{signal-to-interference-plus-noise ratio}
\acrodef{RFID}{radio frequency identification}
\acrodef{WPAN}{wireless personal area network}
\acrodef{WWB}{Weiss-Weinstein bound}
\acrodef{DP}{direct path}
\acrodef{MF}{matched filter}
\acrodef{MMSE}{minimum-mean-square-error}
\acrodef{SBS}{serial backward search}
\acrodef{SBSMC}{serial backward search for multiple clusters}
\acrodef{NBI}{narrowband interference}
\acrodef{WBI}{wideband interference}
\acrodef{INR}{interference-to-noise ratio}
\acrodef{CR}{channel response}
\acrodef{CIR}{channel impulse response}
\acrodef{CR}{channel  response}
\acrodef{RADAR}{radar}
\acrodef{MUR}{Multistatic radar}
\acrodef{JBSF}{jump back and search forward}
\acrodef{HDSA}{high-definition situation-aware}
\acrodef{RRC}{root raised cosine}
\acrodef{ST}{simple thresholding}
\acrodef{BTB}{Bellini-Tartara bound}
\acrodef{P-Max}{$P$-Max}  
\acrodef{MIMO}{multiple-input multiple-output}
\acrodef{MAP}{maximum a posteriori}
\acrodef{FG}{factor graph}
\acrodef{OP}{outage probability}
\acrodef{WED}{wall extra delay}
\acrodef{RMS}{root mean square}
\acrodef{SPAWN}{sum-product algorithm over a wireless network}
\acrodef{MDD}{minimum distance distribution}
\acrodef{MAP}{maximum a posteriori probability}
\acrodef{SAP}{small cell access point}
\acrodef{UE}{user equipment}
\acrodef{MBS}{macro cell base station}
\acrodef{UER}{\ac{UE} Relay}
\acrodef{D2D}{device-to-device}
\acrodef{MBS}{macro base station}
\acrodef{CSI}{channel state information}
\acrodef{OGR}{outage guard region}
\acrodef{FUR}{feasible UER region}
\acrodef{EHR}{energy harvesting region}
\acrodef{EH}{energy harvesting}
\acrodef{D2D-EHSN}{D2D communication provided \ac{EH} small cell network}
\acrodef{D2D-EHHN}{D2D communication provided \ac{EH} heterogeneous network}
\acrodef{3GPP}{3rd Generation Partnership Project}
\acrodef{BS}{base station}
\acrodef{DF}{decode and forward}
\acrodef{CCDF}{complementary cumulative distribution function}
\acrodef{ZF}{zero forcing}
\acrodef{RZF}{regularized zero forcing}
\acrodef{WLLN}{weak law of large number}
\acrodef{SLLN}{strong law of large numbers}
\acrodef{TDD}{Time-division duplex}
\acrodef{EE}{energy efficiency} 
\acrodef{HetNet}{heterogeneous network} 
\acrodef{SCP}{Single Cell Processing}
\acrodef{CBF}{Coordinated Beamforming}
\usepackage{color}
\usepackage{dsfont}
\usepackage{bbm}

\renewcommand{\IEEEQED}{\IEEEQEDopen}

\DeclareMathAlphabet{\mathsf}{OML}{cmbr}{m}{it}

\newtheorem{definition}{\bf Definition}
\newtheorem{theorem}{\bf Theorem}
\newtheorem{lemma}{\bf Lemma}

\newtheorem{remark}{\bf Remark}

\newtheorem{assumption}{\bf Assumption}

\newcommand{\bd}{\begin{description}}
\newcommand{\ed}{\end{description}}
\newcommand{\be}{\begin{enumerate}}
\newcommand{\ee}{\end{enumerate}}
\newcommand{\bi}{\begin{itemize}}
\newcommand{\ei}{\end{itemize}}
\newcommand{\bl}{\begin{list}}
\newcommand{\el}{\end{list}}
\newcommand{\bt}{\begin{tabbing}}
\newcommand{\et}{\end{tabbing}}

\newcommand{\paperTitle}{Robust Federated Learning Over the Air: Combating Heavy-Tailed Noise With Median Anchored Clipping }

\begin{document}

\title{\paperTitle
}

\author{
	\IEEEauthorblockN{
		Jiaxing Li\IEEEauthorrefmark{2}, 
		Zihan Chen\IEEEauthorrefmark{3}, 
		Kai Fong Ernest Chong\IEEEauthorrefmark{3}, 
		Bikramjit Das\IEEEauthorrefmark{3},
		Tony Q. S. Quek\IEEEauthorrefmark{3},
		Howard H. Yang\IEEEauthorrefmark{2}
        } 
	\IEEEauthorblockA{\IEEEauthorrefmark{2}ZJU-UIUC Institute, Zhejiang University, Haining 314400, China}
	\IEEEauthorblockA{\IEEEauthorrefmark{3}Singapore University of Technology and Design, Singapore 487372}
} 


\maketitle
\acresetall
\thispagestyle{empty}
\begin{abstract}
Leveraging over-the-air computations for model aggregation is an effective approach to cope with the communication bottleneck in federated edge learning. 
By exploiting the superposition properties of multi-access channels, this approach facilitates an integrated design of communication and computation, thereby enhancing system privacy while reducing implementation costs.
However, the inherent electromagnetic interference in radio channels often exhibits heavy-tailed distributions, giving rise to exceptionally strong noise in globally aggregated gradients that can significantly deteriorate the training performance. 
To address this issue, we propose a novel gradient clipping method, termed Median Anchored Clipping (MAC), to combat the detrimental effects of heavy-tailed noise. 
We also derive analytical expressions for the convergence rate of model training with analog over-the-air federated learning under MAC, which quantitatively demonstrates the effect of MAC on training performance. 
Extensive experimental results show that the proposed MAC algorithm effectively mitigates the impact of heavy-tailed noise, hence substantially enhancing system robustness. 
\end{abstract}
\acresetall

\section{Introduction}\label{sec:intro} 

In lieu of conventional centralized training methods, federated learning (FL)\cite{MaMMooRam:17AISTATS,Li2019FederatedLC,park2019wireless,liu2022resource} offers a fresh perspective on collaborative data methodologies, enabling clients to benefit from high-quality model services without compromising the security of their private data.
Specifically, clients are not required to upload private local data to the server; rather, they only need to transmit model weights or gradient parameters, effectively averting the exposure of sensitive information. 
FL establishes a collaborative connection between clients and servers, facilitating the joint learning of a shared global model.

Despite the tremendous success of the FL paradigm, there still exist obvious problems while the executing process. 
The frequent transmission of model information between clients and server consumes substantial network bandwidth, while the aggregation of a large number of parameters requires extensive computing resources. 
These factors can result difficulty in accommodating massive access and meeting stringent latency requirements. 
Moreover, although FL avoids the direct aggregation of user data, model parameter information such as gradients can still potentially compromise user privacy. 
Inference attackers may exploit shared gradients to reverse-engineer training data, leading to potential indirect breaches of privacy \cite{lyu2020threats}.

A viable solution to this problem is by integrating over-the-air (OTA) computation\cite{nazer2007computation,yang2020federated,frey2021over,xiao2024over} into the FL system.
By leveraging the superposition property of multiple-access channels, OTA computation enables the automatic aggregation of clients' gradients, significantly improving channel utilization while simultaneously reducing computational overhead \cite{zhu2021over}. 
Furthermore, as the server receives aggregated gradients rather than individual ones from clients \cite{csahin2023survey}, the system's vulnerability to inference attacks is substantially mitigated.

However, the analog channel inherently introduces electromagnetic interference during the transmission \cite{zhu2019broadband,amiri2020machine,guo2020analog,zhang2022turning}.
While such interference enhances privacy protection, it also compromises the reliability of channel transmission, especially when it manifests as impulse interference, rendering the noise exhibiting a heavy-tailed distribution (rather than Gaussian) \cite{chen2023edge}---this has been consistently demonstrated by both theoretical\cite{middleton1977statistical} and empirical evidence \cite{clavier2020experimental}.
In heavy-tailed distributions, extreme values (i.e. very large or very small values) occur with high probability, which could lead to severe signal distortion, resulting in a gradient explosion in the FL system, and thereby profoundly affecting the training process of OTA FL.

Numerous methods have been proposed to combat the impact of strong channel noise, ranging from channel inversion\cite{mital2022bandwidth}, phase correction\cite{sery2020analog,lee2022over}, to amplitude correction and energy estimation\cite{goldenbaum2010computing}.
However, these methods only enhance the channel quality and fail to cope with the heavy tail phenomenon at the algorithmic level.
In the field of deep learning, gradient norm clipping (GNC) has been proposed to solve the gradient explosion problem\cite{pascanu2013difficulty}, 
and has been widely discussed as a solution to the gradient heavy-tailed distribution problem\cite{seetharaman2020autoclip, qian2021understanding, yang2022taming}, but there is a crucial limitation: Once the data statistical structure of the gradient is altered by noise, GNC struggles to maintain its effectiveness.

To enhance the robustness of OTA FL against heavy-tailed noise, we introduce a novel residual clipping technique named median anchored clipping (MAC).
In contrast to GNC, leveraging the stability of the median, this method constrains the magnitude of signals received after centralization, adjusts the proportional relationships among gradients, maximizes gradient retention, and mitigates the impact of heavy-tailed interference on OTA FL.
Our main contributions are summarized as follows:
\begin{itemize}
    \item We propose a novel robust gradient clipping method tailored for OTA FL systems to mitigate the impact of heavy-tailed noise present in the analog channel.
    \item We derive the convergence rate of the OTA FL gradient descent algorithm with MAC under non-convex conditions.
    \item We conducted substantial experiments in which the results show that our MAC algorithm effectively mitigates the impact of heavy-tailed noise in analog OTA FL.
\end{itemize}


\section{System Model}

\subsection{Setting}
\begin{figure}[t!]
\centering
\includegraphics[width=0.47\textwidth]{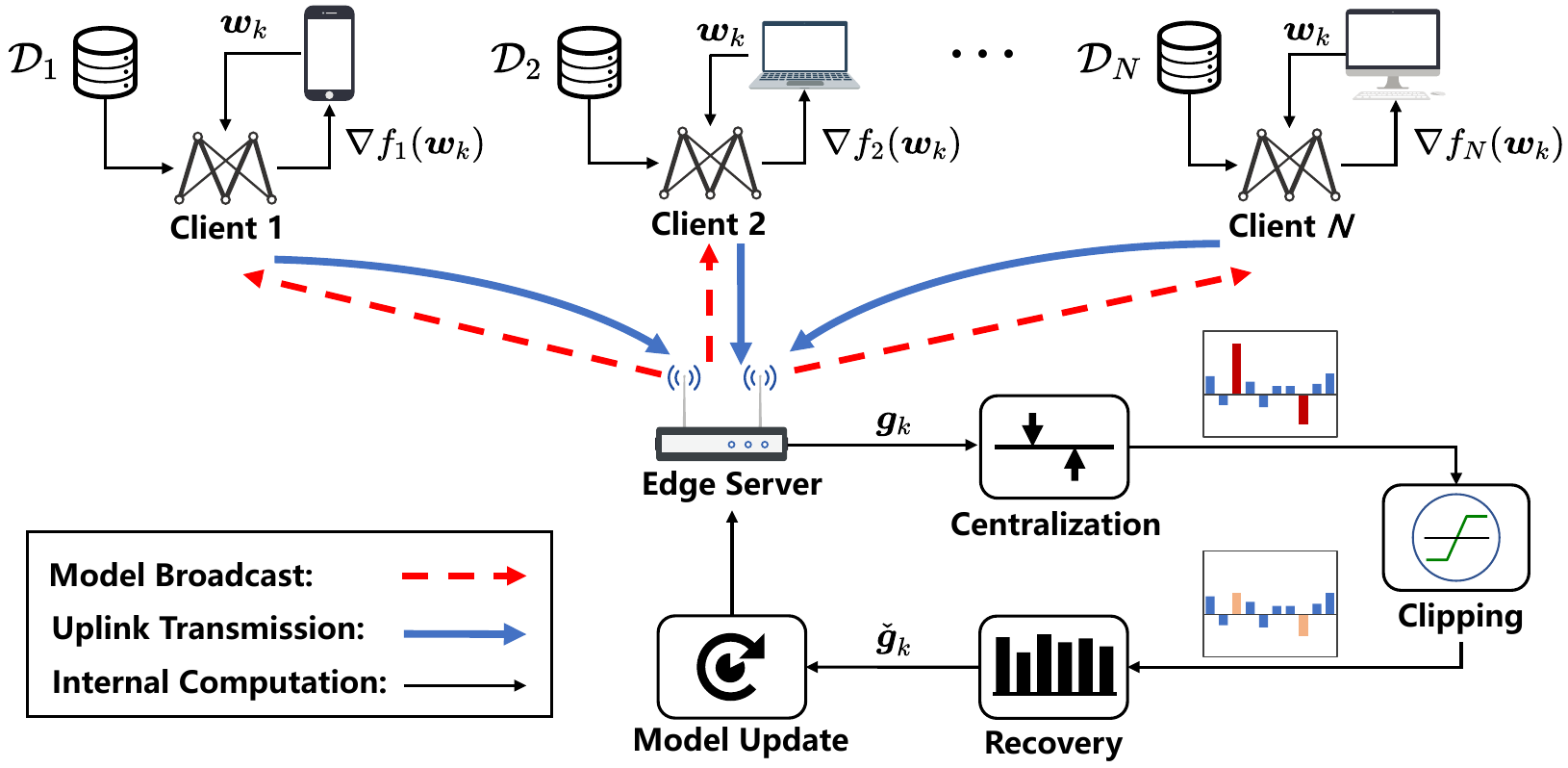} 
\captionsetup{justification=justified, singlelinecheck=false}
\caption{An illustration of the OTA FL training procedure.}
\label{fig:syst_model}
\vspace{-0.5cm}
\end{figure}

We consider the FL system depicted in Fig.~\ref{fig:syst_model}, which consists of an edge server and $N$ clients.
Every client $n$ possesses a local dataset $\mathcal{D}_{n}$ that contains $m_n$ data samples $\{(\boldsymbol{x}_{i}, y_{i} )\}^{m_{n}}_{i=1}$
where $\boldsymbol{x}_{i} \in \mathbb{R}^{d}, y_{i} \in \mathbb{R}$.
We assume that the local datasets are statistically independent from each other.
The edge server orchestrates with the clients to learn a statistical model from their datasets while preserving privacy.

More precisely, the clients need to collaboratively find a vector $\boldsymbol{w} \in \mathbb{R}^d$ that minimizes the following loss function:
\begin{equation}\label{eq:Gbl_loss}
    f(\boldsymbol{w})=\frac{1}{N}\sum^{N}_{n=1}f_{n}(\boldsymbol{w})
\end{equation}
where $f_{n}(\boldsymbol{w})$ is the local empirical risk of agent $n$.
The solution of \eqref{eq:Gbl_loss} is commonly known as the empirical risk minimizer, denoted by
\begin{equation}
    \boldsymbol{w}^{*}=\arg\min f(\boldsymbol{w}).
\end{equation}

In this paper, we adopt the OTA FL for model training, which employs analog transmissions for intermediate parameter uploading, harnessing the superposition property of radio signals for fast gradient aggregation. 

\subsection{Federated Model Training Over the Air}

The general procedure for analog OTA FL is detailed in \cite{yang2021revisiting}. 
We briefly describe it in this part for completeness.
In particular, at the $k$-th round of global communication, the edge server broadcasts the global parameter $\boldsymbol{w}_k$ to all the clients. 
Then, each client $n$ calculates its local gradient $\nabla f_n( \boldsymbol{w}_k )$, modulates this parameter onto the magnitude of a set of common waveforms that are orthogonal to each other, and simultaneously sends the resulting analog signals to the edge server.
The edge server passes the received signal to a bank of matched filters, with each branch tuned to one of the waveform bases, and outputs the automatically aggregated (but distorted) gradient. 
Formally, the global gradient can be written as follows:
\begin{equation}
    \boldsymbol{g}_{k}=\frac{1}{N}\sum_{n=1}^{N}h_{n,k}\nabla f_{n}(\boldsymbol{w}_{k})+\boldsymbol{\xi}_{k}
\end{equation}
in which $h_{n,k}$ represents the channel fading of client $n$ at the $k$-th global iteration, assumed to be a random variable with unit mean and finite variance, independent across the clients, and varies over time in an i.i.d. manner; $\boldsymbol{\xi}_{k}$ results from the electromagnetic interference, modeled as a $d$-dimensional random vector where each entry follows an independent symmetrical $\alpha$-stable distribution ($S\alpha S$)\cite{samorodnitsky1996stable} (with tail index $\alpha$ and scale parameter $\tau$), accounting for the heavy-tailed distribution of impulse noise.  

Consequently, the global parameter is updated as 
\begin{equation} \label{equ:OTA_model_training_vanilla}
    \boldsymbol{w}_{k+1}=\boldsymbol{w}_{k} - \eta\boldsymbol{g}_{k},
\end{equation}
where $\eta$ is the learning rate. 
Then the global parameter will be broadcast to all clients for the next round of computations.

\subsection{Unstable Training Performance}
Normally, the above recursion is executed multiple rounds until convergence (if it occurs), upon which all participating entities have a common model close to $\boldsymbol{w}^*$. 
However, the spectrum is, by nature, a shared medium, giving rise to potentially strong co-channel interference, which typically manifests as noise during training.
A notable feature of analog channel noise is its heavy-tail characteristic\cite{clavier2020experimental}, which is manifested by frequent occurrence of impulse noise, which could lead to gradient explosion.
These impulse noises cause severe distortion in the statistical structure of the gradients, including the proportional relationships between different gradient entries and the distribution of entries with varying numeric magnitudes. Traditional GNC only controls the total norm of the gradient, without addressing the underlying statistical structure. As a result, GNC cannot resolve the noise problem in analog OTA FL.

To that end, the main thrust of the present paper is to develop a scheme to cope with the noise introduced by analog OTA parameter aggregation so as to stabilize the training process and improve the performance of the trained model.

\section{Median Anchored Clipping} 

\begin{figure*}[t!]
\centering
\centering\includegraphics[width=1.0\textwidth]{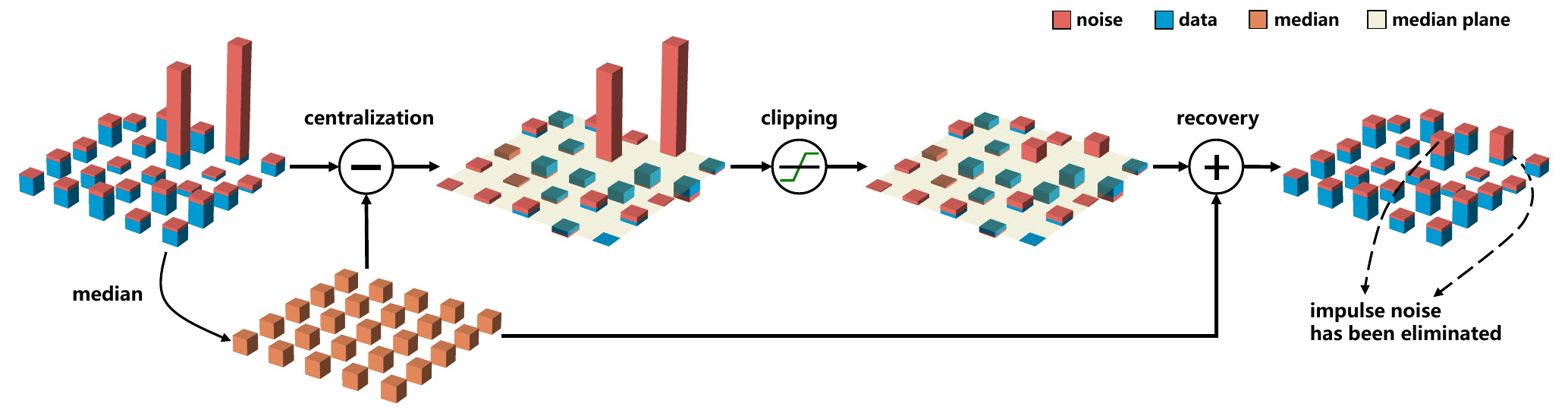} 
\caption{MAC algorithm\textemdash illustrated with a multi-dimensional gradient displayed in a three-dimensional view\textemdash operates as follows: For an aggregated gradient with multiple entries, the vector-median value is subtracted from each entry, clipping is applied, and the gradient is then restored by adding the vector-median value back.}
\label{fig:algorithm diagram}
\vspace{-0.5cm}
\end{figure*}

\subsection{Proposed Method}
As shown in Fig.~\ref{fig:algorithm diagram}, our MAC algorithm strengthens the robustness of OTA FL against strong communication noise by performing three key steps for aggregated gradients: \textit{1) centralization}, \textit{2) clipping}, and \textit{3) recovery}.
The primary concept behind the MAC algorithm is to determine a datum point for a set of gradient entries and, anchoring on this point, recalibrate the magnitudes of entries. 

To begin with, we define the vector-median as follows:
\begin{definition}
    \textit{
    For a vector $\boldsymbol{w} \in \mathbb{R}^{d}$, }\texttt{med}\textit{($\boldsymbol{w}$) is the median entry of all entries of $\boldsymbol{w}$, i.e., given $\boldsymbol{w} = (w_{1}, w_{2}, \dots, w_{d})^\top$,}
    \begin{equation}
        \texttt{med}(\boldsymbol{w})=\mathrm{median}\{w_{i},i \in [d]\}
    \end{equation}
    \textit{where $[d]$ stands for the set $\{1,\dots,d\}$}.
\end{definition}

\noindent\textit{1) Centralization:} 
Given a globally aggregated gradient $\boldsymbol{g}_k$, we centralize it by subtracting the vector-median from each entry, namely,
\begin{equation}
    \boldsymbol{g}_k \leftarrow \boldsymbol{g}_k - \texttt{med}(\boldsymbol{g}_k)\cdot\boldsymbol{1}
\end{equation}
where $\boldsymbol{1}$ represents an all-ones vector.

The rationale behind centralizing the global gradient at the median is that this operation minimizes the $\ell_1$ deviation of the entries (note that due to heavy-tailed noise, the $\ell_2$ deviation of the entries may be unbounded).
As such, during the subsequent clipping procedure, it preserves the original information of entries that are close to the median while eliminates the extreme values introduced by the impulse noise.

\noindent\textit{2) Clipping:}
Based on the centralized gradient $\boldsymbol{g}_k$, we implement value clipping on each entry, thereby restricting the range of individual entries within a designated threshold $C$.
More concretely, for a generic entry $g_{k,i}$, $i \in [d]$, we have
\begin{equation}
    g_{k,i} \leftarrow \texttt{sgn}(g_{k,i})\cdot\min(|g_{k,i}|,C)
\end{equation}
where $\texttt{sgn}(\cdot)$ takes the sign of its input variable. 


\noindent\textit{3) Recovery:}
After clipping the centralized gradient, we add back the median to each entry as follows:
\begin{equation}
    \check{\boldsymbol{g}}_k \leftarrow \boldsymbol{g}_k + \texttt{med}(\boldsymbol{g}_k)\cdot\boldsymbol{1}.
\end{equation}

Ultimately, by employing the MAC algorithm, we obtain a new aggregated global gradient with the detrimental effects of heavy-tailed noise effectively mitigated while retaining the useful information as much as possible.
The details of this method are summarized in Algorithm~\ref{algorithm}.

\begin{algorithm}[tp]
\caption{OTA FL with MAC algorithm}\label{algorithm}
Initialize\  $\boldsymbol{w}_0$

\For{$ k\in[K]$(communication round)}{
    \For{\upshape each\ client\ $n \in [N]$\ in\ parallel}{
        Update\ local\ model: $\boldsymbol{w}_{k,n}=\boldsymbol{w}_{k}$.
        
        Local training: $\nabla f_{n}(\boldsymbol{w}_{k})$.

        Send $\nabla f_{n}(\boldsymbol{w}_{k})$ to the server.
    }
    Global noisy aggregation:
    
        \quad $\boldsymbol{g}_{k} = \frac{1}{N}\sum_{n=1}^{N}h_{k,n}\nabla f_{n}(\boldsymbol{w}_{k})+\boldsymbol{\xi}_{k}$ 

    Server receive $\boldsymbol{g}_{k}$:

        \quad $\Check{\boldsymbol{g}}_{k}=$\textsc{MAC}$(\boldsymbol{g}_{k}, C)$ \textit{// Median Anchored Clipping}

        \quad $\boldsymbol{w}_{k+1}=\boldsymbol{w}_{k} - \eta\check{\boldsymbol{g}}_{k}$ \textit{// Server Update}

        \quad Broadcasting $\boldsymbol{w}_{k+1}$ to clients.
        
}
\algrule
   \SetKwFunction{MAC}{$\mathbf{MAC}$}
    \SetKwProg{Fn}{Function}{:}{}
    \Fn{\MAC{$\boldsymbol{g},\mathit{C}$}}{
    \KwIn{Gradient $\boldsymbol{g}$, threshold $C$}
    \KwOut{Clipped gradient $\boldsymbol{g}$}
    $\boldsymbol{m} = \texttt{med}(\boldsymbol{g}) \cdot \boldsymbol{1}$
    
    $\boldsymbol{g} = \boldsymbol{g} - \boldsymbol{m}$ \qquad \textit{// Centralization} 



    \For{$g_i$ in $\boldsymbol{g}$}
    {
        $g_i = \texttt{sgn}(g_i)\cdot \min\left\{|g_i|, C\right\}$ \quad \textit{// Clipping}
    }

    $\boldsymbol{g} = \boldsymbol{g} + \boldsymbol{m}$ \qquad \textit{// Recovery}
    
    \textbf{return} clipped gradient $\boldsymbol{g}$
}

\end{algorithm}
\setlength{\textfloatsep}{-3pt}

\subsection{Convergence Analysis}
To facilitate the analysis, we make the following assumptions, which are widely adopted in machine learning research. 

\begin{assumption}
    \textit{
    The objective function $f: \mathbb{R}^{d} \rightarrow \mathbb{R}$ is lower bounded by a constant $f(\boldsymbol{w}^*)$, i.e., for any $\boldsymbol{w}\in \mathbb{R}^{d}$, it is satisfied:
\begin{equation}
     f(\boldsymbol{w}) \geq f(\boldsymbol{w}^*)
\end{equation}
    }
\end{assumption}


\begin{assumption}
    \textit{
    The objective function $f: \mathbb{R}^{d} \rightarrow \mathbb{R}$ is $L$-smooth, i.e., for any  $\boldsymbol{w}, \boldsymbol{v} \in \mathbb{R}^{d}$, it is satisfied:
\begin{equation}
     f(\boldsymbol{w}) \leq f(\boldsymbol{v}) + \langle \nabla f(\boldsymbol{v}), \boldsymbol{w}- \boldsymbol{v} \rangle + \frac{L}{2} \Vert \boldsymbol{w} - \boldsymbol{v}\Vert ^2.
\end{equation}
    }
\end{assumption}
\begin{assumption}
    \textit{
    The gradients of each client are bounded, i.e., for $\forall n\in [N]$, there exists a constant $G$ that
    \begin{equation}
        \|\nabla f_n(\boldsymbol{w})\| \leq G.
    \end{equation}
    }
\end{assumption}
At this stage, we are ready to present the main theoretical result of this paper as follows.

\begin{lemma}
    For any gradient entry, let $p_C$ denote the probability it remains unclipped at threshold $C$, there is
    \begin{align}
        p_C  \sim 1 - \frac{\tau^{\alpha}}{C^{\alpha}}.
        \label{p_C}
    \end{align} 
\end{lemma}
\begin{IEEEproof}
Please refer to Appendix B.
\end{IEEEproof}

\begin{theorem}
    \textit{
    If the learning rate is set as $\eta \leq \frac{2}{L}$, and clipping threshold $C > \sqrt{2}G$, then Algorithm \ref{algorithm} converges as
    \begin{align}
        \frac{1}{K}\sum_{k=0}^{K-1}\mathbb{E}&\left[ \|\nabla f(\boldsymbol{w}_{k})\|^2 \right] \leq \frac{2(f(\boldsymbol{w}_{0}) - f(\boldsymbol{w}^*))}{Kp_C(2-\eta L)\eta} \notag \\
        &+\frac{1}{2}\eta^2dL\Big(p_C\big(\frac{\sqrt{2}}{2}C\!-\!G\big)^2 \!+\! (1\!-\!p_C)C^2\Big)
        \label{convergence}
    \end{align}
    }
    \label{throrem1}
\end{theorem}
\begin{IEEEproof}
Please refer to Appendix C.
\end{IEEEproof}
\begin{remark}
    \textit{The result in \eqref{convergence} demonstrates that regardless of the tail index $\alpha$ and scale parameter $\tau$, running OTA FL in conjunction with MAC consistently achieves a sublinear convergence rate (where the residual error can be reduced by decreasing the learning rate). As such, the robustness of model training is substantially enhanced, making it resilient to the detrimental effects of heavy-tailed communication noise. 
    }
\end{remark}
\begin{remark}
    \textit{
        The effects of noise characteristics (including tail index and scale) and clipping threshold are quantified by $p_C$, which is determined within a probabilistic range.
        As \eqref{p_C} shows, these factors jointly influence the algorithm's convergence rate. 
    }
\end{remark}

\begin{figure*}[!htbp]
    \centering
    \subfloat[]
    {
        \includegraphics[width=0.3\textwidth]
    {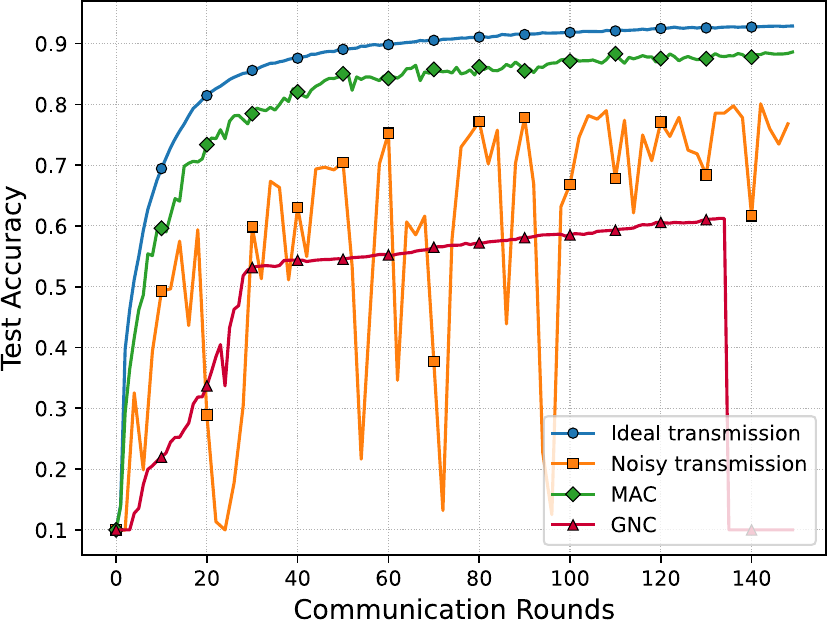}
        \label{fig:subfig_a}
    }
    \subfloat[]
    {
        \includegraphics[width=0.3\textwidth]
    {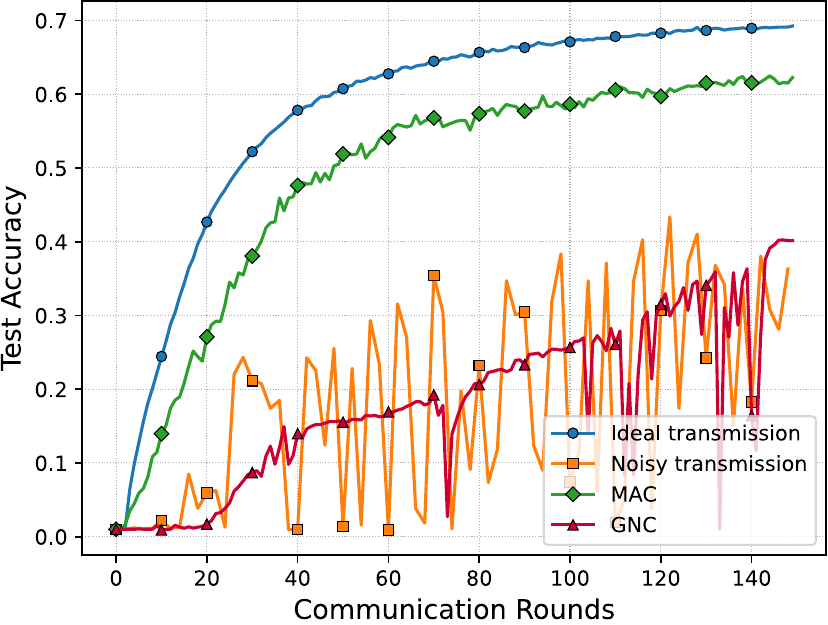}
        \label{fig:subfig_b}
    }
    \subfloat[]
    {
        \includegraphics[width=0.3\textwidth]
    {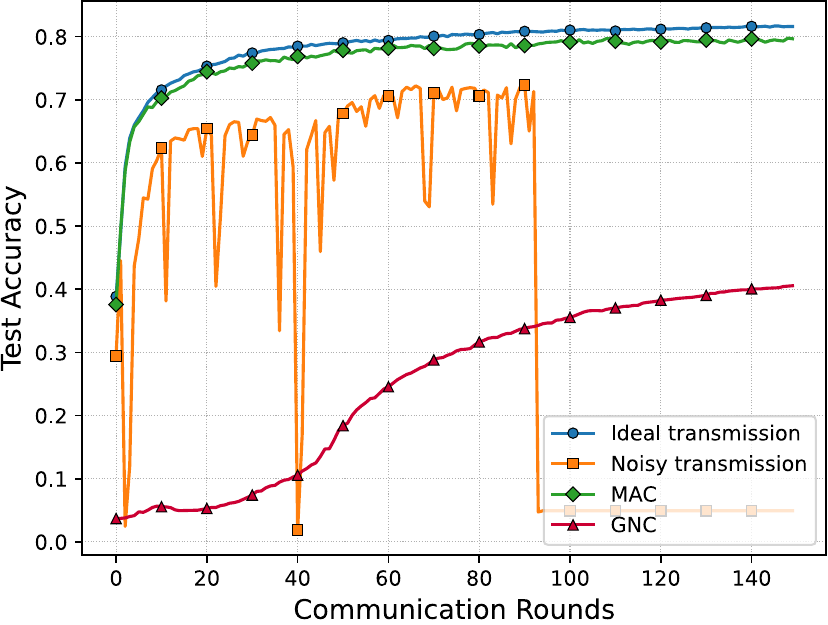}
        \label{fig:subfig_c}
    }

    \subfloat[]
    {
        \includegraphics[width=0.3\textwidth]
    {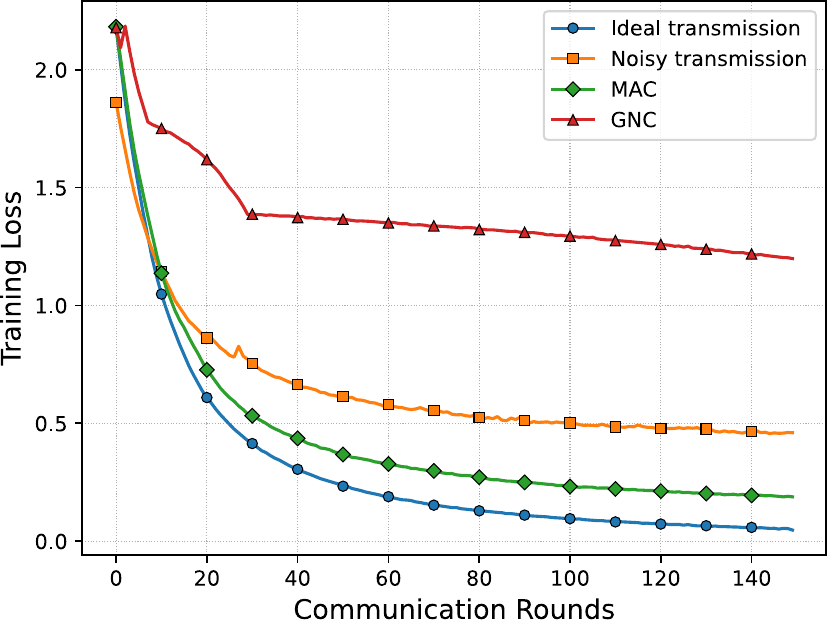}
        \label{fig:subfig_d}
    }
    \subfloat[]
    {
        \includegraphics[width=0.3\textwidth]
    {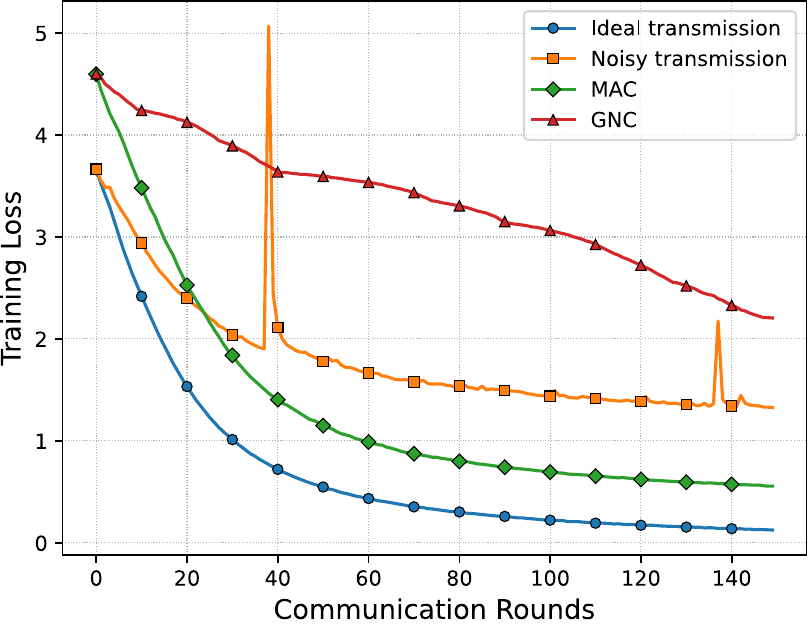}
        \label{fig:subfig_e}
    }
    \subfloat[]
    {
        \includegraphics[width=0.3\textwidth]
    {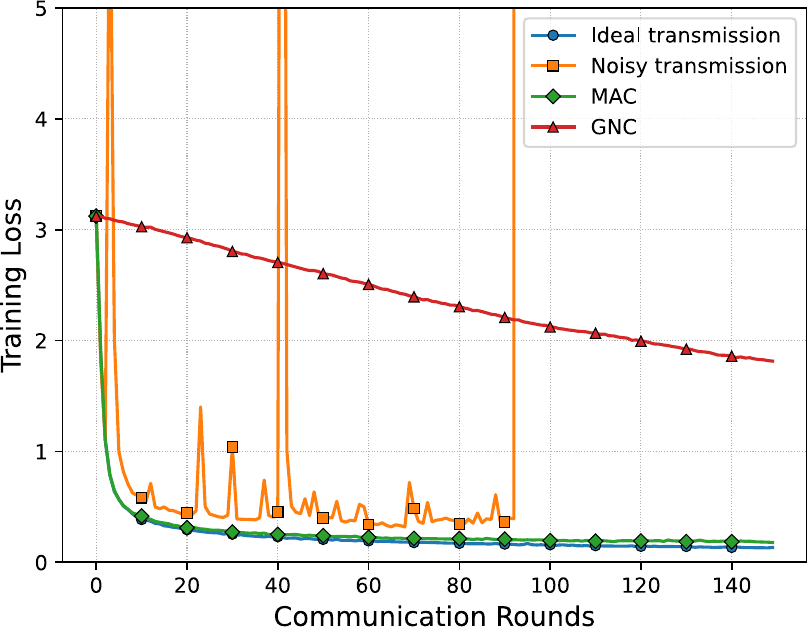}
        \label{fig:subfig_f}
    }
    \caption{Performance comparison between MAC and the baselines, (a) and (d) training ResNet-18 on the CIFAR-10 dataset, (b) and (e) ResNet-34 on the CIFAR-100 dataset, and (c) and (f) training CNN on the FEMNIST dataset.}
    \label{fig:muti-figure1}
    \vspace{-0.5cm}
\end{figure*}

\section{Experimental Results}
\subsection{Experiment Setup}

\textbf{System Setting:}
We evaluated the effectiveness of our proposed MAC algorithm by utilizing the GNC as a baseline and conducted a comparison of their training performance under identical configurations of OTA FL systems. 
Additionally, during the experiment, we performed threshold searches on both the MAC and GNC algorithms and conducted tuning to enhance performance, thereby ensuring the fairness of the comparison.
In our experiment, Rayleigh fading with a parameter setting of $\mu = 1$ is utilized to model channel fading.
Unless otherwise stated, the following parameters will be used: number of clients $N = 50$, learning rate $\eta= 0.03$, local epoch $E = 5$, $S\alpha S$ tail index $\alpha = 1.5$, $S\alpha S$ scale parameter $\tau = 0.1$, and the local batch size is 10.

\textbf{Dataset and Models:}
We assessed performance on CIFAR-10, CIFAR-100\cite{krizhevsky2009learning}, and FEMNIST (which is processed in a non-i.i.d. manner)\cite{caldas2018leaf} using ResNet-18, ResNet-34\cite{he2016deep}, and CNN architectures, respectively. 
As the neural networks have a multi-layer structure, the MAC algorithm is executed parallel to the gradient blocks in a layer-wise manner.
In experiments involving non-independent and identically distributed (non-i.i.d.) data on the CIFAR-10 dataset, we use a Dirichlet distribution with a concentration parameter of $Dir = 0.3$\cite{acar2021federated, hsu2019measuring} to characterize data heterogeneity.





\begin{figure*}[!htbp]
    \centering
    \subfloat[]
    {
        \includegraphics[width=0.3\textwidth]
    {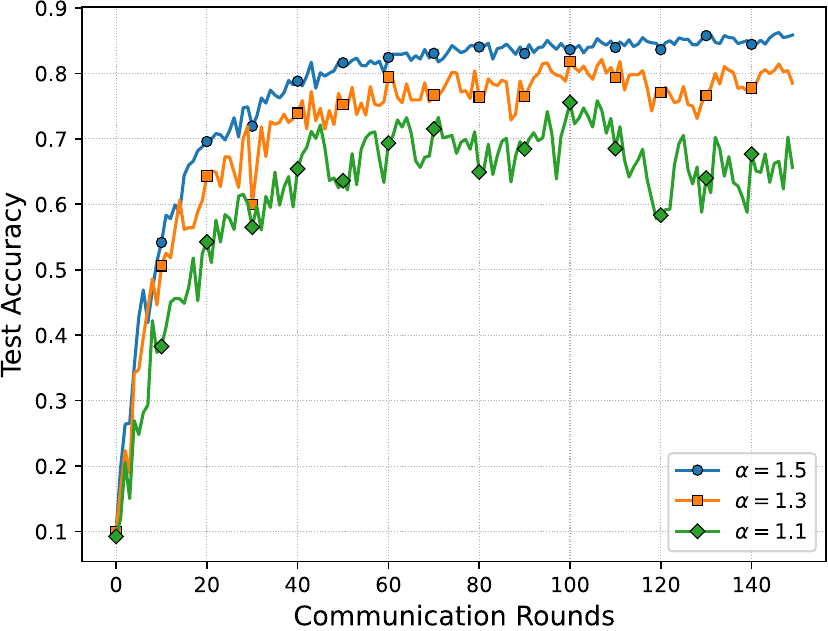}
        \label{fig:subfig_alpha}
    }
    \subfloat[]
    {
        \includegraphics[width=0.3\textwidth]
    {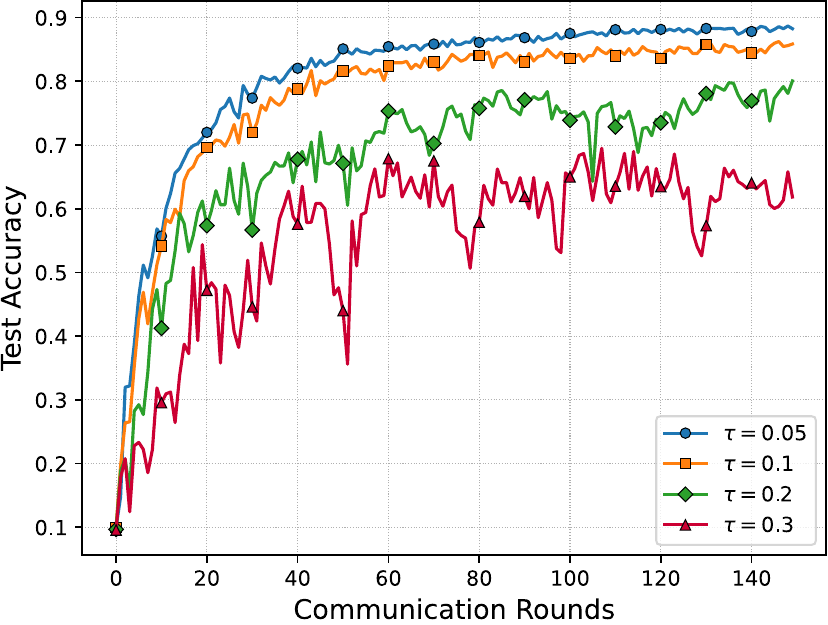}
        \label{fig:subfig_tau}
    }
    \subfloat[]
    {
        \includegraphics[width=0.3\textwidth]
    {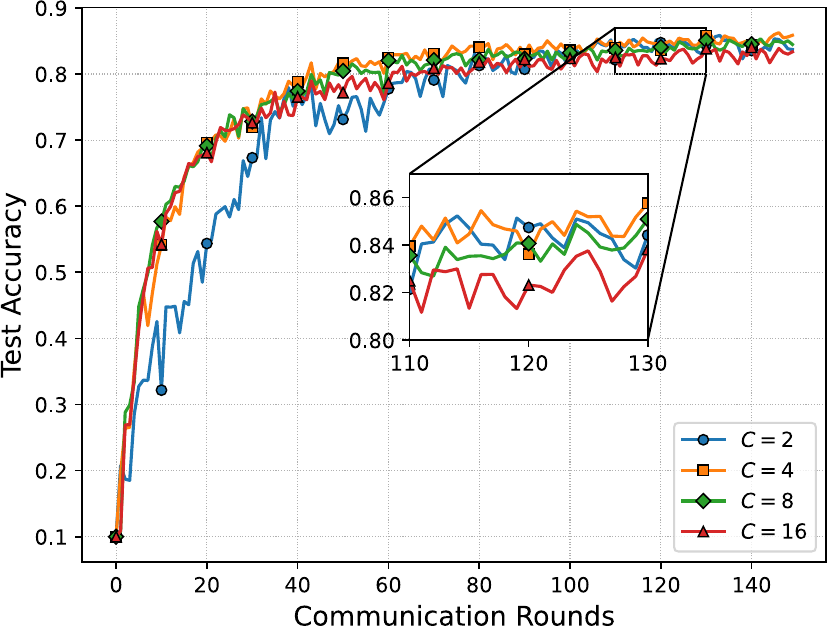}
        \label{fig:subfig_C}
    }
    \caption{Sensitivity evaluation, (a)--(c) are training ResNet-18 on the CIFAR-10 dataset while non-i.i.d. sampling, respectively correspond different tail index $\alpha$, scale parameter $\tau$ and clipping threshold $C$.}
    \label{fig:muti-figure2}
\end{figure*}

\subsection{Performance Evaluation}
In Fig.~\ref{fig:muti-figure1}, we present a comprehensive comparative analysis of the performance of OTA FL across diverse model training tasks.
The automatically aggregated but noisy global gradient gradient, is subjected to various post-processing methodologies: MAC, GNC, and without any post-processing (termed as noisy transmission). 
Furthermore, the ideal scenario is illustrated, wherein channel corruption, encompassing fading and noise, is absent, serving as the upper bound of performance.
The figures presented here demonstrate that the training process is severely impacted by heavy-tailed noise, leading to significant fluctuations in the test accuracy curve, a substantial reduction in the achievable maximum accuracy, and an increase in the loss function value, as shown by the comparison between noisy and ideal transmissions.
In these circumstances, numerous experiments demonstrate that GNC faces difficulties in sustaining effective performance when the statistical structure of gradients, including the proportional relationships among various gradient dimensions and the distribution of entries with differing magnitudes, is significantly disrupted by impulse noise.
Conversely, the application of the MAC algorithm results in a substantial mitigation of these fluctuations, thereby markedly improving test accuracy and training stability, while fortifying the robustness of the OTA FL system.

In Fig.~\ref{fig:muti-figure2}, we examine the performance of the MAC by altering system-related parameters. 
Moreover, distinct from Fig.~\ref{fig:muti-figure1}, all experiments depicted in Fig.~\ref{fig:muti-figure2} are executed under non-i.i.d. conditions, thereby illustrating the universal applicability of our MAC algorithm across datasets characterized by varying degrees of data heterogeneity.
Specifically, Fig.~\ref{fig:muti-figure2}(a) and (b) illustrate a comparative evaluation of the test accuracy of OTA FL MAC across different levels of communication noise.
As shown in Fig.~\ref{fig:muti-figure2}(a), it is evident that decreasing the tail index from $\alpha=1.5$ to $\alpha=1.1$ results in the MAC algorithm maintaining consistent stability while achieving high test accuracy.
Notably, the condition $\alpha=1.1$ represents a scenario with extreme volatility in the channel noise (where $\alpha=1$ corresponds to the Cauchy distribution characterized by undefined mean and variance), whilst MAC demonstrates remarkable robustness even under such extreme noise conditions. 
Fig.~\ref{fig:muti-figure2}(b) examines the performance of the MAC algorithm under conditions of different scale parameter $\tau$. 
It is important to observe that for $\tau = 0.05$, $0.1$, $0.2$, and $0.3$, the Signal-to-Noise Ratios (SNRs), defined as the ratio of useful signal power to channel noise (i.e., SNR$ = \|\nabla f(\boldsymbol{w}_k)\|^{2} / \|\boldsymbol{\xi}_k\|^{2}$), have average values of up to $-35$ dB, $-41$ dB, $-47$ dB, and $-50$ dB, respectively, after measurement.
The figure clearly corroborates that our MAC algorithm exhibits considerable stability in adverse channel conditions, sustaining its performance even when the SNR reaches $-50$ dB.
In alignment with our analysis, Fig.~\ref{fig:muti-figure2}(a) and (b) illustrate that the MAC algorithm substantially mitigates the direct impact of tail index and scale parameters associated with heavy-tailed noise on the convergence rate of OTA FL, thereby enhancing system stability in the extreme case.
On the other hand, Fig. \ref{fig:muti-figure2}(c) presents a comparative analysis of the performance of the MAC algorithm across various clipping thresholds.
The figure reveals that variations in threshold values yield analogous convergence rates, indicating that the MAC algorithm exhibits insensitivity to the clipping thresholds, demonstrating the robustness and ease of parameterization characteristic to the MAC algorithm.

\section{Conclusion}
In this study, we introduce a novel algorithm, denoted as median anchored clipping(MAC), which markedly enhances the robustness of over-the-air federated learning(OTA FL) systems against channel noise, frequently characterized by heavy-tailed distributions.
The MAC algorithm leverages the median of the aggregated global gradient entries as a datum plane and applies value clipping to truncate the extreme values induced by impulse noise, thereby effectively alleviating the detrimental effects of channel noise while largely preserving the original gradient information.
The effectiveness of the MAC algorithm was validated through a convergence analysis conducted under non-convex conditions, alongside a series of empirical experiments that demonstrated the algorithm's consistency across a range of noise conditions. 
The MAC algorithm is effective, low-complex, and straightforward to implement, rendering it highly applicable in practical scenarios.



\bibliographystyle{IEEEtran}
\bibliography{clip_for_ota}
\clearpage

\section{Appendix}
The appendix provides a detailed account of our convergence analysis, which is organized into three subsections: \textit{A. Prerequisites}, \textit{B. Proof of Lemma 1}, and \textit{C. Proof of Theorem 1}. 
Appendix A introduces the analytical framework and establishes the foundational concepts required for the subsequent sections. 
In particular, it outlines a key lemma whose proof is presented in Appendix B. 
Finally, Appendix C concludes with the convergence proof of the MAC algorithm under non-convex conditions.
\subsection{Prerequisites}

First of all, we denote that
\begin{align}
    \nabla f(\boldsymbol{w}_k) = \frac{1}{N}\sum_{n=1}^{N}\nabla f_n(\boldsymbol{w}_{k}),
\end{align}
and since the channel fading among users is assumed to be independently and identically distributed (i.i.d.) with an expected value of 1, we approximate that:
\begin{align}
    \boldsymbol{g}_{k} &= \frac{1}{N}\sum_{i=1}^{N}h_{k,n}\nabla f(\boldsymbol{w}_k) + \boldsymbol{\xi}_k \notag \\
                        &\approx \frac{1}{N}\sum_{n=1}^{N}\nabla f_n(\boldsymbol{w}_k) + \boldsymbol{\xi}_k \notag \\
                        &=\nabla f(\boldsymbol{w}_k) + \boldsymbol{\xi}_k.
\end{align}

In the clipping step of MAC, entries in the aggregated gradient could divided into two parts, some entries are clipped while others are not. 
We could represent this phenomenon by a selection matrix, such as $\boldsymbol{S}_k = \mathrm{diag}\{ s_{k,1}, \dots, s_{k,d} \}$, in which $s_{k,i} \in \{0, 1\}$ indicates whether the $i$-th entry is clipped (in this case, $s_{k,i} = 0$) or not (in this case, $s_{k,i} = 1$).
Consequently, we can express the model update under MAC as follows: 
\begin{align} \label{CLIP update formula}
    &\boldsymbol{w}_{k+1} = \boldsymbol{w}_k - \eta [\boldsymbol{S}_k (\nabla f(\boldsymbol{w_k})+\boldsymbol{\xi}_k) \notag \\
                         &\qquad+ (\boldsymbol{I}-\boldsymbol{S}_k)(\texttt{med}(\nabla f(\boldsymbol{w}_k)+\boldsymbol{\xi}_k)\cdot \boldsymbol{1} + \hat{\boldsymbol{C}}_{k})] \notag \\
                         &= \boldsymbol{w}_k - \eta \big[ \boldsymbol{S}_k \nabla f(\boldsymbol{w}_k) \notag \\
                         &\qquad+ (\boldsymbol{I} - \boldsymbol{S}_k)\cdot\texttt{med}(\nabla f(\boldsymbol{w}_k) + \boldsymbol{\xi}_k)\cdot \boldsymbol{1}\big]  - \eta \boldsymbol{\zeta}_k
\end{align}
where 
$\boldsymbol{I}$ is identity matrix and $\boldsymbol{\zeta}_k$ is 
\begin{align}
    \boldsymbol{\zeta}_k = \boldsymbol{S}_k \boldsymbol{\xi}_k + (\boldsymbol{I} - \boldsymbol{S}_k)\hat{\boldsymbol{C}}_{k}
\end{align}
in which $\hat{\boldsymbol{C}}_{k} = (c_{k,1},\dots, c_{k,d})^{\top}$ where each $c_{k,i}$ is given by 
\begin{align}
    c_{k,i} = \texttt{sgn}(g_{k,i} - \texttt{med}(\nabla f(\boldsymbol{w}_k) + \boldsymbol{\xi}_k)) \cdot C, ~~ i\in [d],
\end{align}
where $\texttt{sgn}(\cdot)$ takes the sign of its input variable.

Notice that the heavy-tailed noise $\boldsymbol{\xi}_k$ is zero-mean, then we make an approximation from median to mean:
\begin{align}
    \texttt{med}(\nabla f(\boldsymbol{w}_k) + \boldsymbol{\xi}_k) \approx  \frac{1}{d}\boldsymbol{1}^\top \nabla f(\boldsymbol{w}_k)
\end{align}

\subsection{Proof of Lemma 1}
Let us start form the range $R$ of entries from gradient $\nabla f(\boldsymbol{w}_k)$, which is defined as
\begin{align}
    R = \text{max}(|\nabla f(w_{k,i}) - \nabla f(w_{k,j})|), \forall i,j \in [d].
\end{align}
Then following the Assumption 3 we know that there has
\begin{align}
    R \leq \sqrt{2}G.
\end{align}
And we assume that $C > \sqrt{2}G$.
Then considering the clipping judge condition, if 
\begin{align}
    |\nabla f(w_{k,i}) + \xi_{k,i} - \texttt{med}(\nabla f(\boldsymbol{w}_k) + \boldsymbol{\xi}_k)| \leq C,
\end{align}
then the i-th entry would not be clipped.
If we denote $\texttt{med}(\nabla f(\boldsymbol{w}_k) + \boldsymbol{\xi}_k)$ as $\nabla f(w_{k,m}) + \xi_{k,m}$, then there has
\begin{align}
    |\nabla f(w_{k,i}) + \xi_{k,i} &- \nabla f(w_{k,m}) - \xi_{k,m}| \notag \\
    &\leq |\nabla f(w_{k,i}) - \nabla f(w_{k,m})| + |\xi_{k,i} - \xi_{k,m}| \notag \\
    &\leq \sqrt{2}G + |\xi_{k,i} - \xi_{k,m}|.
\end{align}
And we denote $\xi_{k,c} = \xi_{k,i} - \xi_{k,m}$, because $\xi_{k,i}$ and $\xi_{k,m}$ are independent and identically distributed as $S\alpha S(\alpha, \tau)$, thus
\begin{align}
    \xi_{k,c} \sim S\alpha S(\alpha, \sqrt{2}\tau).
\end{align}
Then we let
\begin{align}
    \sqrt{2}G + |\xi_{k,c}| \leq C,
\end{align}
which is saying that the clipping will not occur if $|\xi_{k,c}| \leq C-\sqrt{2}G$, the probability is 
\begin{align}
    \mathbb{P}\{|\xi_{k,c}|\leq C-\sqrt{2}G\}.
\end{align}

In a similar way, there the clipping will occur if 
\begin{align}
    C &< |\xi_{k,c}|-|\nabla f(w_{k,i}) - \nabla f(w_{k,m})| \notag \\
      &\leq |\nabla f(w_{k,i}) + \xi_{k,i} - \nabla f(w_{k,m}) - \xi_{k,m}|,
\end{align}
or 
\begin{align}
    C &< |\nabla f(w_{k,i}) - \nabla f(w_{k,m})| - |\xi_{k,c}| \notag \\
      &\leq |\nabla f(w_{k,i}) + \xi_{k,i} - \nabla f(w_{k,m}) - \xi_{k,m}|.
\end{align}
That is 
\begin{align}
    |\xi_{k,c}| > |\nabla f(w_{k, i}) - \nabla f(w_{k, m})| + C
\end{align}
or 
\begin{align}
    |\xi_{k,c}| < |\nabla f(w_{k, i}) - \nabla f(w_{k, m})| - C.
\end{align}
Because $C > \sqrt{2}G$, thus $|\nabla f(w_{k, i}) - \nabla f(w_{k, m})| - C$ is constant negative,
\begin{align}
    \mathbb{P}\{|\xi_{k,c}| < |\nabla f(w_{k, i}) - \nabla f(w_{k, m})| - C\} = 0.
\end{align}
That is when $|\xi_{k,c}| \geq C + \sqrt{2}G$, the clipping will occur, the probability is 
\begin{align}
    \mathbb{P}\{|\xi_{k,c}| > C + \sqrt{2}G\}.
\end{align}

\begin{figure}[hbtp]
    \centering
    \includegraphics[width=1.\linewidth]{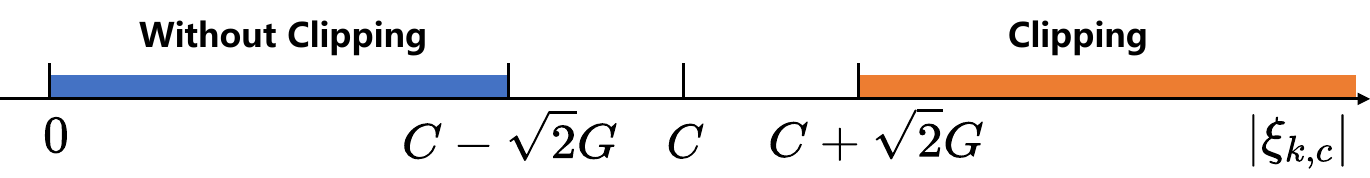}
    \caption{Clipping decision interval}
    \label{fig:Clipping Interval}
\end{figure}
As illustrated in Fig.~\ref{fig:Clipping Interval}, for random variable $|\xi_{k,c}|$, no clipping occurs if $|\xi_{k,c}| \leq C-\sqrt{2}G$.
However, if $|\xi_{k,c}| > C -\sqrt{2}G$, clipping occurs.
In our analysis, we made an approximation wherein this part is entirely subject to clipping.
So, at this stage, we get the probability of without clipping which is denoted as $p_C$, and
\begin{align}
    p_C = \mathbb{P}\{s_i = 1\} &= \mathbb{P}\{|\xi_{k,c}| \leq C -\sqrt{2}G\} \notag \\
                                &= \mathbb{P}\left\{|\xi_{k,i}| \leq \frac{\sqrt{2}}{2}C - G\right\} \sim 1 - \frac{\tau^{\alpha}}{C^{\alpha}}
\end{align}

It is worth noting that our approximation imposes a more stringent clipping mechanism, causing it to occur more frequently.
This, in turn, leads to a more relaxed analytical outcome, which is advantageous compared to the actual scenario being analyzed.
\subsection{Proof of Theorem 1}

We have
\begin{align}
    \label{inequality2}
    \mathbb{E}[f(\boldsymbol{w}_{k+1}) - f(\boldsymbol{w}_k)] &\leq \mathbb{E}[\langle \nabla f(\boldsymbol{w_k}), \boldsymbol{w}_{k+1} - \boldsymbol{w}_k \rangle] \notag \\
    &\qquad\quad + \frac{L}{2}\mathbb{E}[\|\boldsymbol{w}_{k+1} - \boldsymbol{w}_k\|^2]
\end{align}
Then 
\begin{align}
    &\mathbb{E}[\langle \nabla f(\boldsymbol{w}_k), \boldsymbol{w}_{k+1} - \boldsymbol{w}_k \rangle] \notag \\
    &= -\eta \Big (\mathbb{E}[\langle \nabla f(\boldsymbol{w}_k), \boldsymbol{S}_k \nabla f(\boldsymbol{w}_k) \rangle] \notag \\
    &\qquad + \mathbb{E}[\langle \nabla f(\boldsymbol{w}_k),(\boldsymbol{I} - \boldsymbol{S}_k)\texttt{med}(\nabla f(\boldsymbol{w}_k)+ \boldsymbol{\xi}_k)\cdot \boldsymbol{1} \rangle] \notag \\
    &\qquad + \mathbb{E}[\langle \nabla f(\boldsymbol{w}_k), \boldsymbol{\zeta}_k \rangle]\Big) \notag \\
    &\simeq -\eta \Big (\mathbb{E}[\langle \nabla f(\boldsymbol{w}_k), \boldsymbol{S}_k \nabla f(\boldsymbol{w}_k) \rangle] \notag \\
    &\qquad + \mathbb{E}[\langle \nabla f(\boldsymbol{w}_k),(\boldsymbol{I} - \boldsymbol{S}_k)\frac{1}{d}\boldsymbol{1}^{\top}(\nabla f(\boldsymbol{w}_k)+ \boldsymbol{\xi}_k)\cdot \boldsymbol{1} \rangle] \notag \\
    &\qquad + \mathbb{E}[\langle \nabla f(\boldsymbol{w}_k), \boldsymbol{\zeta}_k \rangle]\Big).
\end{align}
Notice that 
\begin{align}
    \mathbb{E}[\boldsymbol{\zeta}_k] = p_C\mathbb{E}[\boldsymbol{\xi}_k] + (1-p_C)\mathbb{E}[\hat{\boldsymbol{C}}_k],
\end{align}
and $\mathbb{E}[\boldsymbol{\xi}_k]$ is at the condition $|\xi_{k,i}|\leq \frac{\sqrt{2}}{2}C - G$, where $i \in [d]$.
For vector $\hat{\boldsymbol{C}}_{k}$, due to the properties of the median, the following holds probabilistically for $i \in [d]$:
\begin{align}
    \mathbb{P}\{c_{k,i} = C\} = \mathbb{P}\{c_{k,i} = -C\} = 0.5,
\end{align}
so we have
\begin{align}
    \mathbb{E}[\boldsymbol{\zeta}_k] = \boldsymbol{0}.
\end{align}
Then we get 
\begin{align}
    \label{inequality1}
    &\mathbb{E}[\langle \nabla f(\boldsymbol{w}_k), \boldsymbol{w}_{k+1} - \boldsymbol{w}_k \rangle] \notag \\
    &\simeq -\eta \Big( p_C\mathbb{E}\left[ \|\nabla f(\boldsymbol{w}_k)\|^2 \right] \!+\! \frac{1}{d}(1\!-\!p_C) \mathbb{E}\left[ \|\boldsymbol{1}^{\top}\nabla f(\boldsymbol{w}_k)\|^2 \right] \Big).
\end{align}
Notice that $\boldsymbol{S}_{k}^{\top} \boldsymbol{S}_{k} = \boldsymbol{S}_{k}$, we have
\begin{align}
    \label{inequality3}
    &\mathbb{E}\left[ \|\boldsymbol{w}_{k+1} - \boldsymbol{w}_k\| \right] \notag \\
    &\simeq \eta^2 \mathbb{E}\bigg[ \Big\|\boldsymbol{S}_k\nabla f(\boldsymbol{w}_k) + (\boldsymbol{I}-\boldsymbol{S}_k)\frac{1}{d}\boldsymbol{1}^{\top}\nabla f(\boldsymbol{w}_k)\cdot \boldsymbol{1} + \boldsymbol{\zeta}_k \bigg\|^2 \bigg] \notag \\
    &=\eta^2 \Big[ \mathbb{E}\big[ \|\boldsymbol{S}_k \nabla f(\boldsymbol{w}_k)\|^2 \big] \!+\! \frac{1}{d^2}\mathbb{E}\big[ \|(\boldsymbol{I}-\boldsymbol{S}_k)\boldsymbol{1}^{\top}\nabla f(\boldsymbol{w}_k) \!\cdot\! \boldsymbol{1} \|^2 \big] \notag \\
    &\qquad \qquad + \mathbb{E}\big[\|\boldsymbol{\zeta}_k\|^2\big]\Big]\notag \\
    &=\eta^2 \Big( p_C\mathbb{E}\big[ \|\nabla f(\boldsymbol{w}_k)\|^2\big] \!+\! \frac{1}{d}(1-p_C)\mathbb{E}\big[ \|\boldsymbol{1}^{\top}\nabla f(\boldsymbol{w}_k)\|^2\big] \notag \\
    &\qquad \qquad + \mathbb{E}\big[ \|\boldsymbol{\zeta}_k\|^2 \big]\Big).
\end{align}
Then from inequalities (\ref{inequality1}), (\ref{inequality2}) and (\ref{inequality3}), we get that 
\begin{align}
    \mathbb{E}[f(\boldsymbol{w}_{k+1}) &- f(\boldsymbol{w}_k)] \leq -\left(1-\frac{1}{2}\eta L\right)\eta p_C \mathbb{E}\left[\|\nabla f(\boldsymbol{w}_k)\|^2\right] \notag \\
    &+\left( \frac{1}{2}\eta L - 1 \right)\eta (1 - p_C) \mathbb{E}\left[ \|\boldsymbol{1}^{\top}\nabla f(\boldsymbol{w}_k)\|^2 \right] \notag \\
    &+ \frac{1}{2}\eta^2 L \mathbb{E}\left[ \|\boldsymbol{\zeta}_k\|^2 \right].
\end{align}
Let $\eta$ be constant, and satisfies that $\eta \leq \frac{2}{L}$, then we get
\begin{align}
    \mathbb{E}[f(\boldsymbol{w}_{k+1}) - f(\boldsymbol{w}_k)] &\leq -\left(1-\frac{1}{2}\eta L\right)\eta p_C \mathbb{E}\left[\|\nabla f(\boldsymbol{w}_k)\|^2\right] \notag \\
    &\qquad + \frac{1}{2}\eta^2 L \mathbb{E}\left[ \|\boldsymbol{\zeta}_k\|^2 \right].
\end{align}
For $\mathbb{E}\left[ \| \boldsymbol{\zeta}_k \|^2 \right]$, notice that $s_{k,i}^{2} = s_{k,i}$, we have
\begin{align}
    \mathbb{E}\left[\|\boldsymbol{\zeta}_k\|^2\right] &= \mathbb{E}\left[ \left\| \boldsymbol{S}_k\boldsymbol{\xi}_k + (\boldsymbol{I} - \boldsymbol{S}_k)\hat{\boldsymbol{C}}_k \right\|^2 \right] \notag \\
    &=\mathbb{E}\bigg[ \Big\|\boldsymbol{S}_k \boldsymbol{\xi}_k \Big\|^2 \bigg] + \mathbb{E}\left[ \left\|(\boldsymbol{I}-\boldsymbol{S}_k)\hat{\boldsymbol{C}}_k \right\|^2 \right] \notag \\
    &=\mathbb{E}\left[ \sum_{i=1}^{d}s_{k,i}^{2}\xi_{k,i}^2\right] + \mathbb{E}\left[ \sum_{i=1}^{d}(1 - s_{k,i})^2c_{k,i}^2 \right] \notag \\
    &=\mathbb{E}\left[ \sum_{i=1}^{d}s_{k,i}\xi_{k,i}^2\right] + \mathbb{E}\left[ \sum_{i=1}^{d}(1 - s_{k,i})c_{k,i}^2 \right] \notag \\
    &=d p_C \mathbb{E}\left[ \xi_{k,i}^2 \Bigg| |\xi_{k,i}| \leq \frac{\sqrt{2}}{2}C\!-\!G \right] \!+\! d(1\!-\!p_C)C^2 \notag \\
    &\leq d p_C \left(\frac{\sqrt{2}}{2}C-G \right)^2 + d(1 - p_C)C^2.
\end{align}
Then we get that
\begin{align}
    \left ( 1-\frac{1}{2}\eta L \right)&\eta p_C \mathbb{E}\left[ \|\nabla f(\boldsymbol{w}_k)\|^2 \right] \leq \mathbb{E}[f(\boldsymbol{w}_k)] - \mathbb{E}[f(\boldsymbol{w}_{k+1})] \notag \\
    & + \frac{1}{2}\eta^2 dL\left[  p_C \left(\frac{\sqrt{2}}{2}C\!-\!G\right)^2 \!+\! (1\!-\!p_C)C^2\right].
    \label{iterate}
\end{align}
Summing inequality (\ref{iterate}) for all $k \in \{0,1,...,K-1\}$, and rearranging the results we get 
\begin{align}
    \frac{1}{K}\sum_{k=0}^{K-1}\mathbb{E}&\left[ \| \nabla f(\boldsymbol{w}_k)\|^2 \right] \leq \frac{2(f(\boldsymbol{w}_0)-\mathbb{E}[f(\boldsymbol{w}_{K})])}{Kp_C(2-\eta L)\eta} \notag \\
    &+ \frac{1}{2}\eta^2 d L\left[  p_C \left(\frac{\sqrt{2}}{2}C-G\right)^2 + (1-p_C)C^2\right]
\end{align}
which completes the proof.



\end{document}